\documentclass{article}
\usepackage{iclr2027_conference,times}

\usepackage{amsmath,amsfonts,bm}

\def\eqref#1{equation~\ref{#1}}

\def\1{\bm{1}}

\DeclareMathAlphabet{\mathsfit}{\encodingdefault}{\sfdefault}{m}{sl}
\SetMathAlphabet{\mathsfit}{bold}{\encodingdefault}{\sfdefault}{bx}{n}

\usepackage{url}
\usepackage{booktabs}
\usepackage{colortbl}
\usepackage{graphicx}
\usepackage{pgfplots}
\usepgfplotslibrary{groupplots}
\pgfplotsset{compat=1.18}
\usepackage{float}
\usepackage{hyperref}
\usepackage{needspace}
\title{TutorialGUI: Towards Generalizable GUI Agents via Tutorial-Driven Skill Learning}

\author{
  Shuxuan Fu\textsuperscript{1,2},
  Qi Yao\textsuperscript{3},
  Gaorong Liang\textsuperscript{1,2},
  Jianqiang Huang\textsuperscript{1,2,*},
  Baisheng Lai\textsuperscript{1,2,*}\\[4pt]
  \normalfont\small\textsuperscript{1}Computer Network Information Center, Chinese Academy of Sciences\\
  \normalfont\small\textsuperscript{2}University of Chinese Academy of Sciences\\
  \normalfont\small\textsuperscript{3}Alibaba Group\\[2pt]
  \normalfont\small\textsuperscript{*}Corresponding authors.
}

\hypersetup{
  pdftitle={TutorialGUI: Towards Generalizable GUI Agents via Tutorial-Driven Skill Learning},
  pdfauthor={Shuxuan Fu, Qi Yao, Gaorong Liang, Jianqiang Huang, Baisheng Lai}
}

\iclrfinalcopy
\begin{document}

\maketitle
\pagestyle{plain}

\begin{abstract}
GUI agents have made rapid progress in operating general-purpose software, yet they struggle to generalize to unfamiliar specialized applications due to limited application-specific knowledge and operational experience. Collecting data and fine-tuning agents for each application is costly and difficult to scale to long-tail applications. Official tutorials provide a readily available source of domain knowledge and procedural guidance, but they are human-oriented and cover only a subset of application capabilities. To address these limitations, we introduce \textbf{TutorialGUI}, a training-free framework that transforms official tutorials into executable, reusable skills for GUI agents through an iterative \textbf{Replay--Exploration} loop. Starting from tutorial-derived candidate skills, Replay completes and refines their procedures through real GUI execution and verifies whether they achieve the intended effects. Exploration builds on verified anchor skills to discover new candidate skills via three exploration directions: forward, backward, and lateral. The Replay-Exploration loop can run multiple iterations, progressively extending skills beyond the original tutorials. The resulting skills and associated Knowledge Cards can be retrieved and reused by different GUI agents without updating the model parameters. On six ScienceBoard applications, multimodal TutorialGUI improves task success by an average of 19.45 percentage points across three executor models over the same models without tutorial resources (Vanilla). These results support tutorial-driven skill learning as an effective approach to improving GUI agents' generalization to specialized applications without per-application fine-tuning.

\end{abstract}

\section{Introduction}
\label{sec:introduction}

\suppressfloats[t]
\begin{figure}[t]
    \centering
    \includegraphics[width=\linewidth]{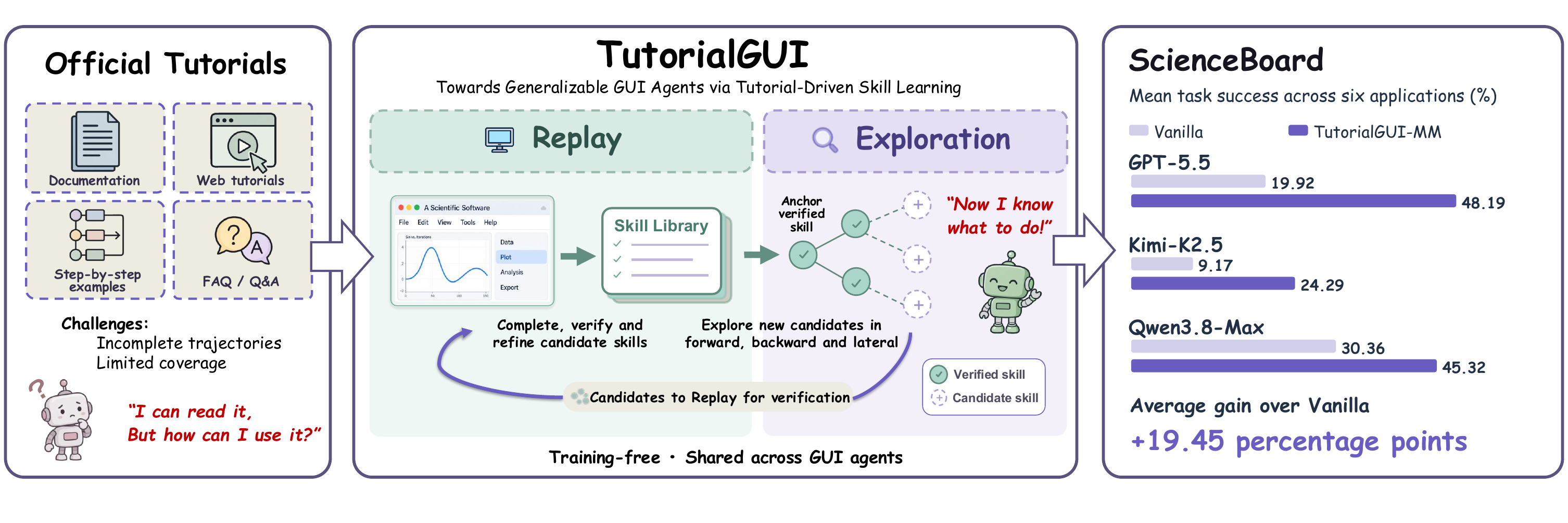}
    \caption{TutorialGUI turns official tutorials into reusable skills through Replay and Exploration, improving GUI agents' performance on scientific software without parameter updates.}
    \label{fig:tutorialgui-overview}
\end{figure}

GUI agents have made rapid progress in interacting with general-purpose applications. Native models such as UI-TARS \citep{qin2025uitars} improve perception, reasoning, and action generation through large-scale interaction data and iterative training, while benchmarks such as OSWorld \citep{xie2024osworld} support evaluation in real desktop environments. Despite this progress, unfamiliar specialized applications remain challenging: agents may lack the application-specific knowledge needed to select tools, configure parameters, and follow specialized workflows \citep{xie2026guide,sun2026scienceboard}. Application-specific training can help address this gap \citep{sun2025coda}, but repeatedly collecting data and updating models introduces adaptation and maintenance costs, limiting scalability to long-tail applications. This motivates complementary approaches that acquire application knowledge and operational experience without per-application fine-tuning.

Scientific software provides a concrete setting for studying this challenge. The tasks in these applications require understanding domain-specific objects, parameter semantics, and operational preconditions while executing multi-step procedures with strict dependencies \citep{sun2026scienceboard}. Successful execution therefore requires both conceptual understanding and procedural knowledge: agents must understand what application-specific concepts mean, when particular operations are appropriate, and how to carry them out in the correct sequence. These complementary forms of knowledge are important for reliable interaction with unfamiliar specialized applications.

Official tutorials offer a readily available source of such knowledge through functional explanations and procedural examples. Existing work has explored synthesizing training trajectories from web tutorials \citep{xu2025agenttrek} and extracting video demonstrations for inference-time guidance \citep{iu2025onlinevideos}. Turning tutorials into reusable skills poses two challenges: executability and coverage. First, \textbf{ensuring executability} requires more than extracting a sequence of steps. Tutorials written for human users may omit intermediate actions or assume prerequisite states; the extracted procedures therefore need to be completed and verified in the actual application. Second, \textbf{extending coverage} requires discovering operations beyond those demonstrated in tutorials. Since tutorials typically cover selected workflows, extraction alone leaves related operations and alternative procedures unexplored.

To address these challenges, we introduce \textbf{TutorialGUI}, a training-free framework that organizes tutorial-derived procedures and application knowledge into a shared external library. It extracts executable and extended skills describing operational procedures and Knowledge Cards providing static knowledge, such as tool functionality and parameter semantics. At the core of skill acquisition is a \textbf{Replay--Exploration loop} that improves skill executability and expands skill coverage. Replay uses real GUI execution to complete missing steps, refine candidate procedures, and verify their intended effects. Exploration uses verified skills as anchors to explore new candidate skills in forward, backward and lateral directions. The Replay-Exploration loop can run multiple iterations, progressively extending skills beyond the original tutorials. During task execution, different GUI agents retrieve relevant skills and Knowledge Cards as contextual augmentation, without parameter updates.

On scientific software tasks in ScienceBoard \citep{sun2026scienceboard}, multimodal TutorialGUI improves task success rates by an average of \textbf{19.45 percentage points} across multiple models over the corresponding Vanilla models. These results demonstrate the effectiveness of the overall framework in supporting scientific software use through tutorial-driven knowledge and skill acquisition without model fine-tuning.

Our main contributions are as follows:
\begin{itemize}
\item \textbf{We introduce TutorialGUI}, a training-free framework that transforms official tutorials into reusable operational skills and Knowledge Cards capturing non-procedural application knowledge. These resources form a shared external library reusable across GUI agents to improve generalization to unfamiliar specialized applications without per-application fine-tuning.
\item \textbf{We propose a Replay-Exploration loop to improve skill executability and expand skill coverage.} Replay completes, refines, and verifies candidate procedures through real GUI execution, while Exploration builds on verified skills to explore in forward, backward and lateral directions to discover new skills beyond the original tutorials.
\item \textbf{We evaluated TutorialGUI across multiple open- and closed-source models on ScienceBoard}, with the multimodal configuration improving average task success by 19.45 percentage points over the corresponding Vanilla models.
\end{itemize}

\section{Related Work}
\label{sec:related_work}

\subsection{GUI Agents}

GUI agents aim to accomplish user goals by interpreting interface observations and executing sequences of actions. Benchmarks including OSWorld~\citep{xie2024osworld,yuan2026osworld2}, Windows Agent Arena~\citep{bonatti2024windows}, and MacArena~\citep{muryn2026macarena} evaluate this ability in real desktop environments, while models such as UI-TARS~\citep{qin2025uitars,wang2025ui} and Mobile-Agents~\citep{xu2026mobileagent35} advance GUI perception and action generation. Successful task execution requires these capabilities to be coordinated across a workflow: agents must select appropriate operations, track changes in application state, and determine how intermediate results support subsequent steps.

\subsection{GUI Agents for Scientific Software}

Operating scientific software requires agents to understand domain-specific tools, parameter semantics, and workflow dependencies. ScienceBoard~\citep{sun2026scienceboard} provides a benchmark for evaluating agents on such scientific workflows. Training-based approaches acquire application-specific capabilities from interaction data~\citep{fan2025guibee,liuosexpert2026,liu2026scalecua}. For example, SEAgent~\citep{sun2025seagent} combines autonomous exploration with reinforcement learning, while CODA~\citep{sun2025coda} trains scientific-software planners through reinforcement learning and supervised fine-tuning. TutorialGUI instead uses official tutorials and GUI interaction to build a reusable external library of application knowledge and operational skills, keeping model parameters fixed.

\subsection{Learning from Tutorials for GUI Agents}

Software tutorials provide procedural knowledge and demonstrations that GUI agents can learn from. One line of work converts these resources into training supervision. AgentTrek \citep{xu2025agenttrek} extracts tasks and instructions from web tutorials and generates training trajectories through guided replay in real environments. Subsequent work scales tutorial-based supervision by identifying interaction steps and aligning them with screen observations to construct large-scale training corpora across diverse applications \citep{zhang2026tongui,lu2026videoagenttrek,xiong2026video2gui}. \mbox{OS-Wizard}~\citep{liu2026oswizard} uses tutorials to guide branching from saved interface states, collecting trajectories of GUI actions and application-native commands for supervised fine-tuning.

Another line of work uses tutorial-derived knowledge as external context during task execution. GUIDE \citep{xie2026guide} extracts planning and grounding knowledge from tutorial videos, while other methods convert videos into demonstration trajectories that agents consult during execution \citep{iu2025onlinevideos,song2026watch}. WebLearner \citep{long2026guiding} further trains agents to retrieve and apply multimodal tutorials. RAG-PolicyJudge~\citep{nguyen2026tutorials} conditions both its action proposals and the judge's choice among them on a task-specific multimodal tutorial.

Several recent approaches also extract skills from instructional resources, yet they differ notably from TutorialGUI. TutorialGUI builds around a Replay–Exploration loop: it refines individual skills and continually expands the skill library through interaction with the real environment. In contrast, MMG2Skill~\citep{che2026mmg2skill} revises guide-derived skills across repeated attempts at the same task, without a separate stage for exploring new operations. VisualSkill~\citep{jiang2026visualskil} uses free and trajectory-targeted UI exploration to enrich its guides with screenshots, without requiring a separate replay check for each procedure. Resource2Skill~\citep{fan2026resource2skill} expands its library by searching for additional source material and distilling new skills from it, rather than discovering skills through in-environment interaction.

\section{Method}

\label{sec:method}

\begin{figure}[!t]
    \centering
    \includegraphics[width=\linewidth]{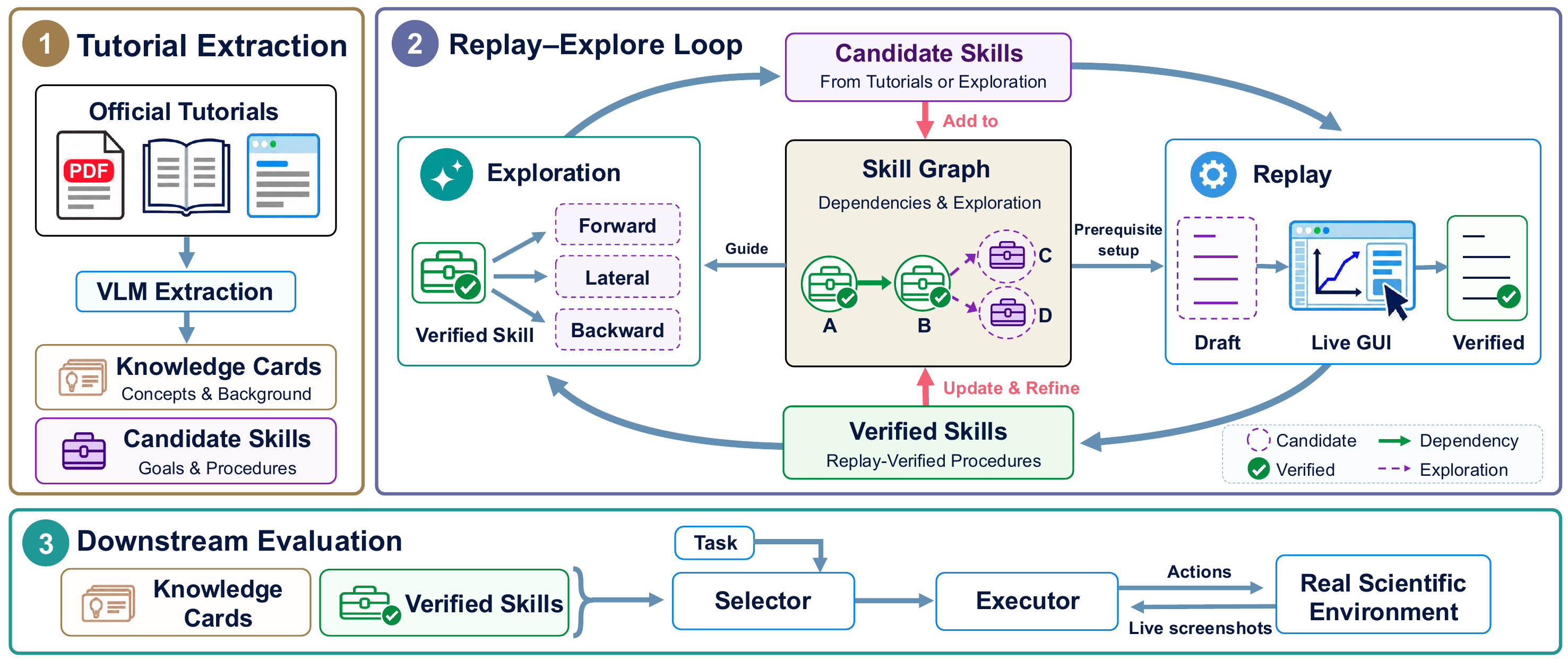}
    \caption{Overview of TutorialGUI. Official tutorials provide candidate skills and Knowledge Cards. Replay refines and verifies candidates through GUI execution, while Exploration proposes related candidates from verified skills. The resulting resources are frozen before evaluation and selected as context for downstream agents without parameter updates.}
    \label{fig:tutorialgui-pipeline}
\end{figure}

TutorialGUI extracts candidate skills and Knowledge Cards from official tutorials. Its Replay--Exploration loop refines and verifies candidate procedures through GUI interaction, then uses verified skills in the Skill Graph to discover further candidates. The resulting library provides retrievable context for downstream agents. Figure~\ref{fig:tutorialgui-pipeline} provides an overview.

\subsection{Tutorial-Based Skill and Knowledge Extraction}

\label{sec:skill_extraction}

Given an application's official tutorials, a vision-language model (VLM) reads their text and images subsection by subsection. It extracts operational procedures as candidate skills and complementary knowledge as Knowledge Cards, organizing the candidate skills into a graph.

\Needspace{11\baselineskip}
\textbf{Candidate skills.} The skills defined in this work are anchored to an operational goal. Operational procedures for different goals are extracted as distinct skills. We formulate a skill as $s=\langle d,P,\Theta,A,Q\rangle$,
where $d$ describes the goal, $P$ specifies preconditions that may be established by other skills, $\Theta$ defines the parameter schema for reuse with different inputs, $A$ describes the operational procedure, and $Q$ specifies expected effects. For example, a KAlgebra plotting skill can bind its expression parameter to $\ln(x)$ or $\sin(2x)$, reusing the procedure for different inputs. We prompt a VLM to extract these fields from tutorials, yielding an initial candidate set $\mathcal{S}_0=\{s_i\}$. Replay further refines these candidates to resolve omitted operational details and ambiguous preconditions.

\textbf{Skill Graph.} We organize relationships among skills in a continuously updated Skill Graph. Nodes represent individual skills, and edges capture two types of relationships: execution dependencies and exploration relations. Execution dependency edges are determined by skill preconditions and guide the execution order during Replay. Exploration edges link newly discovered candidates to their anchor skills, recording how the library expands through Exploration.

\textbf{Knowledge Cards.} Beyond the operational procedures for each function, a GUI agent requires essential background knowledge to operate the application correctly. We therefore extract non-executable knowledge records from tutorials to form Knowledge Cards, which contain software-related concepts, functional explanations, applicability conditions, and interface knowledge. Knowledge Cards can also be retrieved to augment the context of the GUI agent, and remain fixed after extraction.

\subsection{Replay: Execution, Verification, and Refinement}

\label{sec:replay}

Replay establishes whether a candidate procedure can achieve its intended effect and uses the resulting execution evidence to refine it. Guided by dependency relations in the Skill Graph, Replay selects a candidate skill $s_i=\langle d_i,P_i,\Theta_i,A_i,Q_i\rangle$. A dependency identifies a procedure for establishing a required precondition; it does not imply that the current GUI state already satisfies that precondition.

The Execution Agent first checks $P_i$ against the current application state in the environment $\mathcal{E}$ and performs preparation operations when necessary. It instantiates the parameters in $\Theta_i$ using the replay inputs and current application state, yielding bindings $b_i$, then executes $A_i$ once the required preconditions and parameter values are established.

\Needspace{8\baselineskip}
Replay returns the bindings $b_i$ and execution trace $\tau_i$, including preparation, actions, screenshots, and any executor-supplied file probes; a VLM-based postcondition verifier $V$ checks the declared effects under these bindings against the execution evidence:
\begin{align}
    (b_i,\tau_i)&=\operatorname{Replay}(s_i,\mathcal{E}),\\
    v_i&=\mathbf{1}\!\left[V(Q_i,b_i,\tau_i)=\texttt{verified}\right]\in\{0,1\}.
\end{align}

The verifier $V$ returns \texttt{verified}, \texttt{failed}, or \texttt{inconclusive}; the indicator is one only when all required postconditions pass. For admission, $(b_i,\tau_i)$ refers to the first verified attempt, or the final unsuccessful attempt if the Replay budget is exhausted. Admission verifies the observed instance. Appendix~\ref{app:parameter_verification} tests whether the verification label stays the same when a skill is executed with new parameter values.

After successful verification, refinement records the procedure $A_i^\star$ and preconditions $P_i^\star$ supported by that trajectory as $s_i^\star=\langle d_i,P_i^\star,\Theta_i,A_i^\star,Q_i\rangle$. Direct Replay of tutorial-derived candidates yields $\mathcal{S}_0^\star=\{s_i^\star\mid s_i\in\mathcal{S}_0,\ v_i=1\}$. Candidates with no verified attempt within the Replay budget are not admitted and receive no further Replay attempts in subsequent rounds.

\subsection{Exploration: Iterative Skill Expansion}

\label{sec:explore}

Another challenge in extracting skills from tutorials is incomplete coverage: tutorials typically cover only selected workflows, leaving many functions and procedures uncovered. Accordingly, we propose the Exploration module to discover skills beyond the scope of tutorials. Starting from a verified anchor skill \(s_i^\star\), Exploration considers three directions:

\begin{itemize}
\item Forward exploration identifies downstream operations that consume an object produced by the anchor or a state established by it.
\item Backward exploration identifies procedures that establish the anchor's required inputs or preconditions, including missing or alternative preparation paths.
\item Lateral exploration identifies operations with shared inputs, related goals, or neighboring functions, including alternative modes or configurations within the same skill family, without implying an execution dependency.
\end{itemize}

Figure~\ref{fig:explore-structure-example} illustrates these relations in ChimeraX: starting from an anchor that sets the surface style to \texttt{Solid}, forward exploration considers adjusting the surface transparency, while lateral exploration considers the alternative \texttt{Dot} surface style.
\begingroup
\let\originalfigure\figure
\renewcommand{\figure}[1][]{\originalfigure[!t]}
\begin{figure}[!htbp]
    \centering
    \edef\explorepanelheight{\the\dimexpr 0.265\textwidth\relax}
    \begin{minipage}[t]{0.325\textwidth}
        \vspace{0pt}\centering
        \includegraphics[height=\explorepanelheight, width=\linewidth,keepaspectratio]{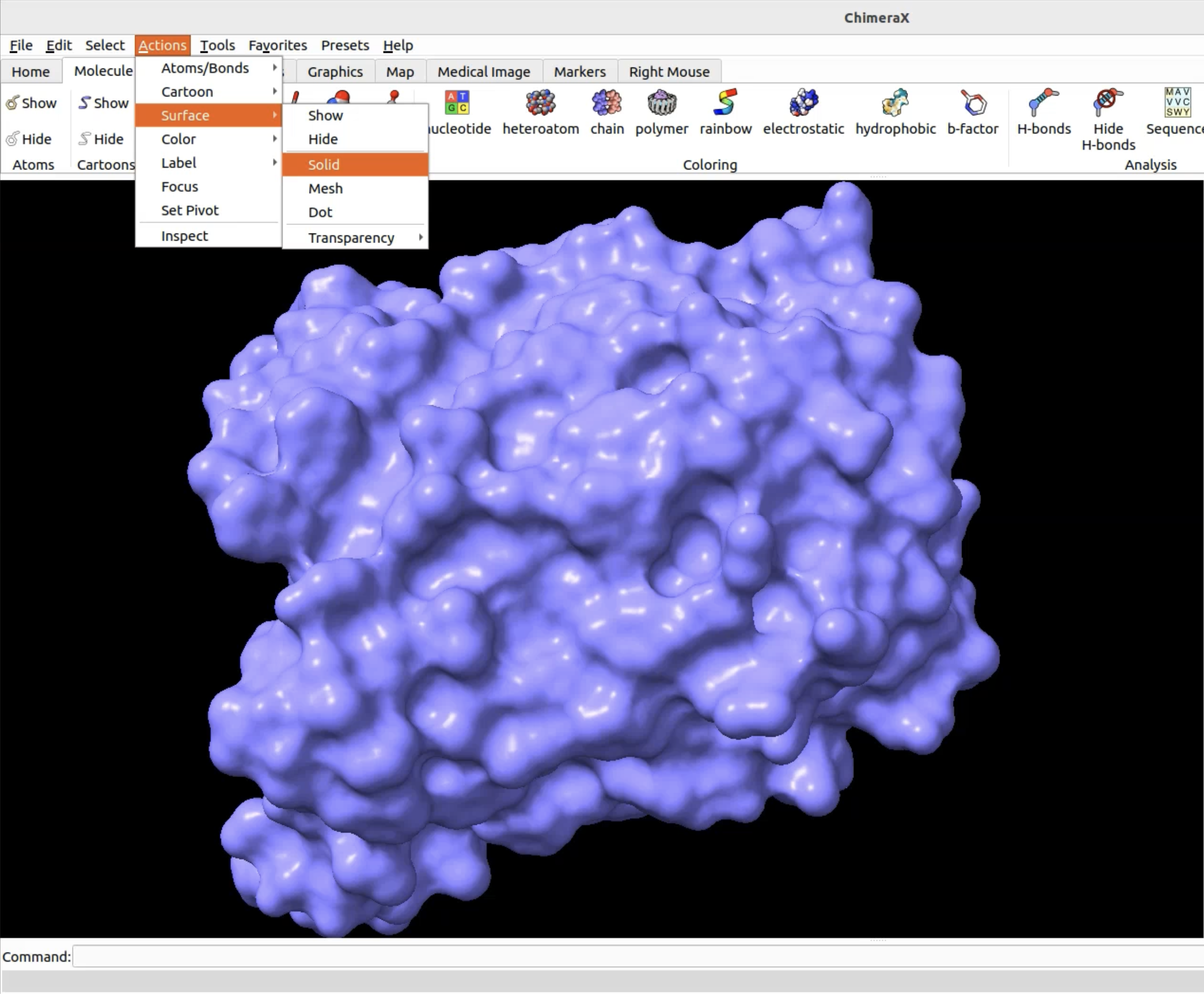}\par
        \smallskip
        {\footnotesize\textbf{(a) Anchor}\\Set surface style to Solid\par}
    \end{minipage}\hfill
    \begin{minipage}[t]{0.325\textwidth}
        \vspace{0pt}\centering
        \includegraphics[height=\explorepanelheight, width=\linewidth,keepaspectratio]{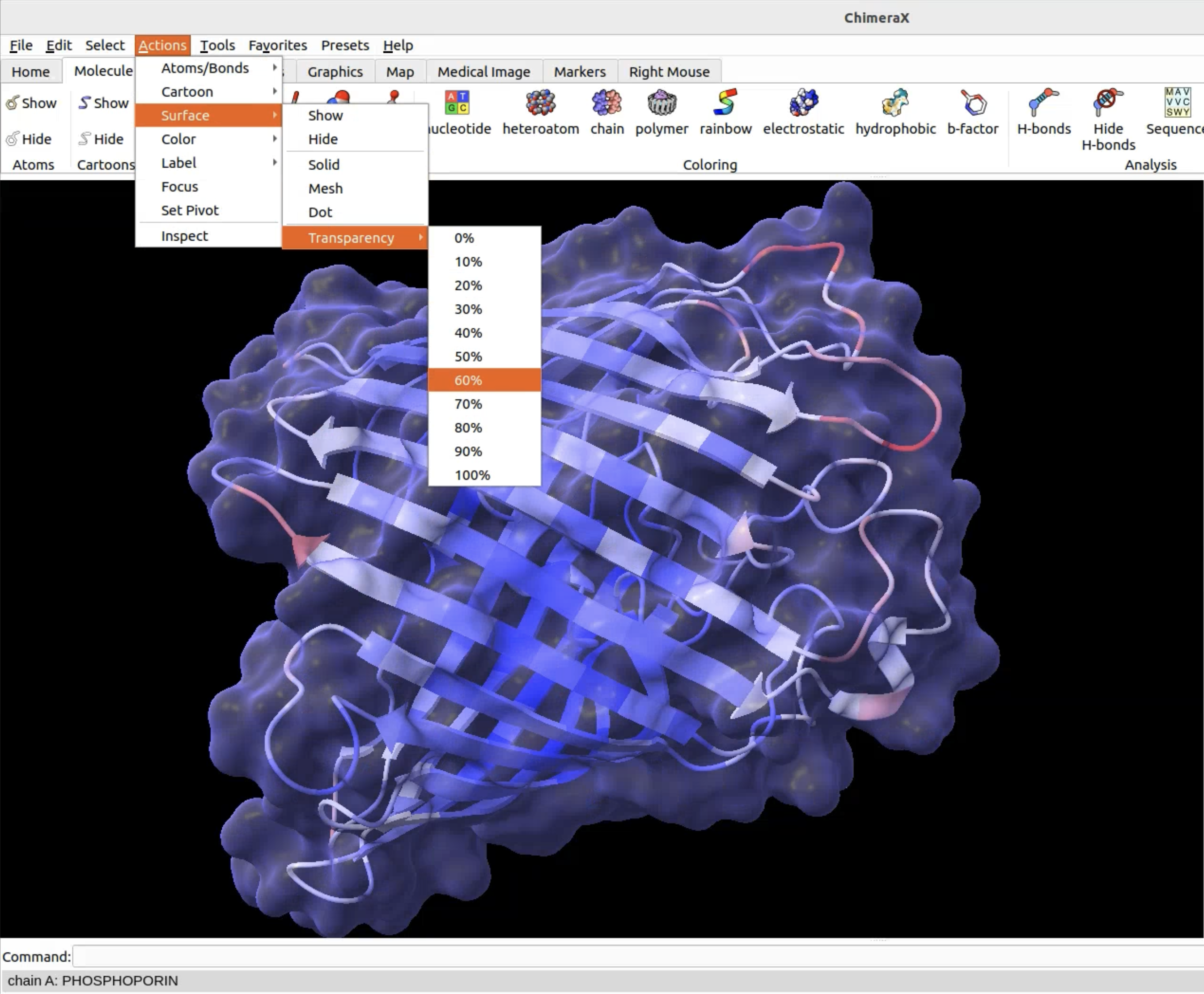}\par
        \smallskip
        {\footnotesize\textbf{(b) Forward}\\Adjust surface transparency\par}
    \end{minipage}\hfill
    \begin{minipage}[t]{0.325\textwidth}
        \vspace{0pt}\centering
        \includegraphics[height=\explorepanelheight, width=\linewidth,keepaspectratio]{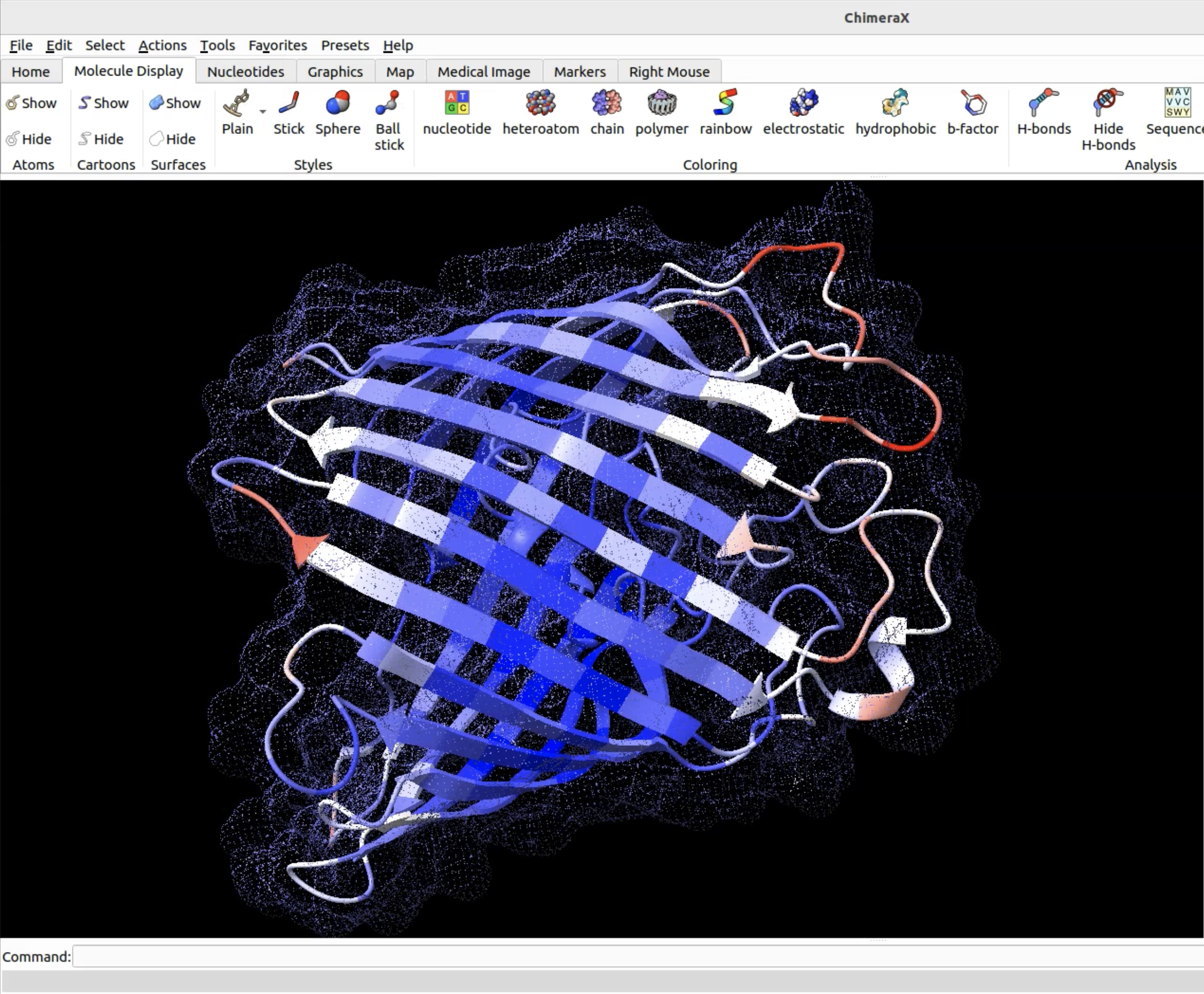}\par
        \smallskip
        {\footnotesize\textbf{(c) Lateral}\\Set surface style to Dot\par}
    \end{minipage}
    \caption{Exploration from a ChimeraX surface-style anchor: (a) the anchor sets the style to \texttt{Solid}; (b) forward exploration considers adjusting surface transparency, with available settings shown; (c) lateral exploration considers the alternative \texttt{Dot} style.}
    \label{fig:explore-structure-example}
\end{figure}

\endgroup

When selecting anchor skills for exploration, verified skills that have not yet been used as exploration anchors are prioritized. Exploration proposes candidates from these anchors, application observations, and relevant tutorial or manual evidence. Candidates remain unverified until Replay (Appendix~\ref{app:exploration_prompt}).

\Needspace{8\baselineskip}
For round $t\geq 1$, let $\mathcal{S}_{t-1}^{\star}$ and
$\mathcal{S}_{t}^{\star}$ denote the verified skill sets at the
start and end of the round, respectively, and let $\mathcal{G}$
denote the current Skill Graph. Candidate generation and the
round-level verified-set update are summarized as
\begin{align}
    \mathcal{U}_t
    &= \operatorname{Explore}(\mathcal{G},\mathcal{E}),\\
    \mathcal{S}_t^{\star}
    &= \mathcal{S}_{t-1}^{\star}\cup
    \left\{s_i^{\star}\mid s_i\in\mathcal{U}_t,\ v_i=1\right\}.
\end{align}
Here, $\operatorname{Explore}$ summarizes successive
candidate-generation calls interleaved with Replay, and
$\mathcal{U}_t$ collects all proposals made during the round.
Only candidates that pass verification are refined and admitted
as new records. The graph is updated upon admission, making newly
verified skills available for subsequent exploration within the
same round. Previously verified skills are retained unchanged
and are not replayed as part of this update. A round ends when the library admits the target number of new Replay-verified skills for the application. If Exploration can no longer produce a new skill that passes Replay, expansion stops for that application.

\subsection{Skill-Augmented Task Execution}

\label{sec:skill_augmented_execution}

After resource construction, the resulting skills augment downstream GUI agents.

\textbf{Selector.} Given the task instruction, the Selector first examines skill summaries and Knowledge Card topics and summaries to identify applicable resources. It selects at most two skills and one Knowledge Card, which are provided to the Executor as task context. If no resources are suitable, the Selector abstains. The Executor binds skill parameters using the task instruction and current GUI observations.

\textbf{Executor.} We use a standard ReAct agent loop for task execution. Each evaluated VLM serves as the Executor, using the selected skills and Knowledge Cards as additional context to guide GUI actions.

\textbf{Text and Multimodal Skills.} TutorialGUI supports text and multimodal variants. For Replay-verified skills, the multimodal variant supplements the textual record with one final-state image from the verified trajectory to limit token overhead. Unverified candidates use original tutorial images when available.

\section{Experiments}
\label{sec:experiments}

We evaluate whether the same tutorial-derived resources help three different GUI agents complete scientific tasks.
We then use GPT-5.5 to examine which resources account for the gains and how task success changes with each round of Exploration.

\subsection{Experimental Setup}
\label{sec:experimental_setup}

We evaluate 169 ScienceBoard tasks~\citep{sun2026scienceboard} spanning algebra (KAlgebra), biochemistry (ChimeraX), geographic information systems (GRASS GIS), automated theorem proving (Lean~4 in VS Code), astronomy (Celestia), and scientific documentation (TeXstudio).
We run the applications in AgentBay sandboxes\footnote{\url{https://www.alibabacloud.com/help/en/agentbay}} on Ubuntu 22.04 at $1920 \times 1080$, using ScienceBoard's software configurations and official task evaluators.
For each method we run, we average success over three independent runs per task, then give each application equal weight in the overall mean.
We report differences in percentage points (pp). Appendix~\ref{app:run_variability} gives the variation across runs, and Table~\ref{tab:task_skill_counts} gives the number of tasks per application.

We run GPT-5.5~\citep{openai2026gpt55}, Kimi-K2.5~\citep{team2026kimi}, and Qwen3.8-Max~\citep{qwen2026qwen38} first without tutorial resources (Vanilla) and then with the same frozen library of Knowledge Cards and skills.
In TutorialGUI-Text, the model receives the selected resources as text. TutorialGUI-MM also supplies one reference image for each selected skill when available.
Both variants observe live GUI screenshots and use the same model to select resources and execute actions (Appendix~\ref{app:experimental_details}).

We evaluate MMG2Skill~\citep{che2026mmg2skill} with all three models in the same environment. With GPT-5.5, we also evaluate Tutorial RAG (retrieval over the source tutorials), VisualSkill~\citep{jiang2026visualskil}, and Resource2Skill~\citep{fan2026resource2skill}.
Table~\ref{tab:main_results} includes published scores for three trained GUI agents, CODA Stage-2~\citep{sun2025coda}, SEAgent with UI-TARS-1.5-7B-DPO~\citep{sun2025seagent}, and ScaleCUA with Qwen3.5-9B~\citep{lv2026scalecua}. Appendix~\ref{app:comparison_protocol} details their application coverage and evaluation protocols.

\subsection{Main Results}
\label{sec:main_results}
\textbf{TutorialGUI achieves substantial gains over Vanilla across all three model families.}
Table~\ref{tab:main_results} shows that the text resources alone raise mean success by 20.94 pp for GPT-5.5, 12.61 pp for Kimi-K2.5, and 11.23 pp for Qwen3.8-Max over their respective Vanilla scores.
With reference images, these gains reach 28.27, 15.12, and 14.96 pp, respectively. Most of the improvement for each executor thus comes from the resources available in the text configuration.
The largest improvement over Vanilla comes on KAlgebra for GPT-5.5 (12.90\% to 80.65\%) and Qwen3.8-Max (41.94\% to 74.19\%). Kimi-K2.5 gains most on ChimeraX (29.89\% to 52.87\%).

\begin{table}[!t]
    \centering
    \caption{
    Task success on ScienceBoard (\%). Within each foundation-model block, bold marks the highest score in each application and in Avg, including ties. Avg weights the reported applications equally. Vanilla uses no tutorial resources. $\Delta$ is the change in Avg from Vanilla (pp). $^{\dagger}$Published results follow their original protocols.
    }
    \label{tab:main_results}
    \setlength{\tabcolsep}{3pt}
    \renewcommand{\arraystretch}{1.12}
    \resizebox{\linewidth}{!}{%
    \begin{tabular}{lrrrrrrrr}
        \toprule
        Method & Alg & Biochem & GIS & ATP & Astron & Doc & Avg & $\Delta$ \\
        \midrule
        \rowcolor[gray]{0.92}
        \multicolumn{9}{l}{\textit{Published trained GUI agents}} \\
        CODA Stage-2$^{\dagger}$
        & 20.16 & 32.23 & 14.71 & --- & 17.05 & --- & 21.04 & --- \\
        SEAgent (UI-TARS-1.5-7B-DPO)$^{\dagger}$
        & 21.29 & 23.45 & 10.59 & --- & 11.52 & --- & 16.71 & --- \\
        ScaleCUA (Qwen3.5-9B)$^{\dagger}$
        & 67.7 & 62.1 & 35.3 & 19.0 & 60.6 & 86.7 & 55.23 & --- \\
        \midrule
\rowcolor[gray]{0.92}
\multicolumn{9}{l}{\textit{GPT-5.5}} \\
Vanilla & 12.90 & 44.83 & 12.75 & 3.85 & 20.20 & 25.00 & 19.92 & --- \\
Tutorial RAG & 40.86 & 51.72 & 8.82 & 3.85 & 21.21 & 20.83 & 24.55 & +4.63 \\
MMG2Skill & 48.39 & 50.57 & 22.55 & 3.85 & 19.19 & 4.17 & 24.79 & +4.87 \\
VisualSkill & 52.69 & 58.62 & 17.65 & 5.13 & 27.27 & 10.42 & 28.63 & +8.71 \\
Resource2Skill & 29.03 & 56.32 & 23.53 & 3.85 & 25.25 & 10.42 & 24.73 & +4.81 \\
TutorialGUI-Text & 67.74 & 74.71 & 40.20 & 5.13 & 30.30 & 27.08 & 40.86 & +20.94 \\
TutorialGUI-MM & \textbf{80.65} & \textbf{86.21} & \textbf{49.02} & \textbf{7.69} & \textbf{34.34} & \textbf{31.25} & \textbf{48.19} & +28.27 \\
\midrule
\rowcolor[gray]{0.92}
\multicolumn{9}{l}{\textit{Kimi-K2.5}} \\
Vanilla & 10.75 & 29.89 & 0.98 & 0.00 & 5.05 & 8.33 & 9.17 & --- \\
MMG2Skill & \textbf{32.26} & \textbf{52.87} & 9.80 & 3.85 & 22.22 & 12.50 & 22.25 & +13.08 \\
TutorialGUI-Text & 29.03 & 40.23 & \textbf{14.71} & \textbf{7.69} & 18.18 & 20.83 & 21.78 & +12.61 \\
TutorialGUI-MM & 29.03 & \textbf{52.87} & 11.76 & 3.85 & \textbf{23.23} & \textbf{25.00} & \textbf{24.29} & +15.12 \\
\midrule
\rowcolor[gray]{0.92}
\multicolumn{9}{l}{\textit{Qwen3.8-Max}} \\
Vanilla & 41.94 & 60.92 & \textbf{17.65} & 10.26 & 22.22 & 29.17 & 30.36 & --- \\
MMG2Skill & 55.91 & 58.62 & \textbf{17.65} & 11.54 & 31.31 & 37.50 & 35.42 & +5.06 \\
TutorialGUI-Text & \textbf{74.19} & 78.16 & 14.71 & \textbf{12.82} & 38.38 & 31.25 & 41.59 & +11.23 \\
TutorialGUI-MM & \textbf{74.19} & \textbf{82.76} & 14.71 & 8.97 & \textbf{45.45} & \textbf{45.83} & \textbf{45.32} & +14.96 \\
        \bottomrule
    \end{tabular}%
    }
\end{table}

\textbf{TutorialGUI leads the tutorial-based comparisons on GPT-5.5.}
As shown in Table~\ref{tab:main_results}, Tutorial RAG raises mean success from 19.92\% to 24.55\%. MMG2Skill and Resource2Skill obtain similar means (24.79\% and 24.73\%), and VisualSkill reaches 28.63\%.
TutorialGUI already exceeds these means with text alone (40.86\%). Its multimodal version reaches 48.19\% and has the highest success rate among the tutorial-based methods on each of the six applications.
With Kimi-K2.5 and Qwen3.8-Max, TutorialGUI-MM also achieves higher mean success than MMG2Skill (24.29\% versus 22.25\% and 45.32\% versus 35.42\%, respectively).
On all four applications for which CODA Stage-2~\citep{sun2025coda} and SEAgent (UI-TARS-1.5-7B-DPO)~\citep{sun2025seagent} report results, TutorialGUI-MM with GPT-5.5 also scores higher than both agents under their respective evaluation protocols.
We attribute part of TutorialGUI's advantage to its Replay--Exploration loop. Replay refines individual procedures through execution in the application, and Exploration uses verified procedures to discover additional skills. The stage-wise ablation in Table~\ref{tab:stage_ablation} examines the contribution of each stage.

Gains over Vanilla vary across applications. On GRASS GIS, TutorialGUI-MM raises success from 12.75\% to 49.02\%, compared with 23.53\% for Resource2Skill, the strongest tutorial-based comparison on that application. On Lean~4, success rises from 3.85\% to only 7.69\%.
Differences in task difficulty, the executor's capabilities in each domain, and the quality of the available tutorials may all contribute to this variation.

\textbf{The value of a reference image depends on the application.}
Adding images to the selected text resources improves the six-application mean most for GPT-5.5 (+7.33 pp), followed by Qwen3.8-Max (+3.73 pp) and Kimi-K2.5 (+2.51 pp).
On ChimeraX, all three executors benefit from images. On KAlgebra, GPT-5.5 gains from images, while Kimi-K2.5 and Qwen3.8-Max attain the same scores with text alone.
On Lean~4, both Kimi-K2.5 and Qwen3.8-Max do better with text alone, and Qwen3.8-Max with images falls below its Vanilla score (8.97\% versus 10.26\%).
In total, reference images raise success in 12 of the 18 model--application pairs, leave it unchanged in three, and reduce it in three.

\subsection{Ablation Studies}
With GPT-5.5, we examine how Replay and Exploration contribute to the skill library, what Knowledge Cards and skills each contribute at execution, and how success changes with the number of Exploration rounds.

\subsubsection{Stage-wise Ablation of Replay and Exploration}
\label{sec:stage_ablation}
Table~\ref{tab:stage_ablation} follows the same executor through four settings. In the Vanilla setting, the executor acts without tutorial resources. Candidate adds the Knowledge Cards and the procedures extracted directly from tutorials, with tutorial images where available. Direct Replay replaces these candidates with procedures that have been executed, refined, and verified in the GUI, together with reference images from those executions. The final setting also includes skills discovered through Exploration and admitted by Replay. The Cards and downstream execution budget are fixed across the three tutorial-based settings.

\begin{table}[!t]
\centering
\caption{Stage-wise ablation of Replay and Exploration with GPT-5.5. Entries are task success rates (\%). The three tutorial-based settings use the same Knowledge Cards.}
\label{tab:stage_ablation}
\small
\setlength{\tabcolsep}{4pt}
\renewcommand{\arraystretch}{1.10}
\begin{tabular}{lrrrrrrr}
\toprule
Construction stage & Alg & Biochem & GIS & ATP & Astron & Doc & \textbf{Avg} \\
\midrule
Vanilla & 12.90 & 44.83 & 12.75 & 3.85 & 20.20 & 25.00 & \textbf{19.92} \\
Candidate & 56.99 & 49.43 & 6.86 & 3.85 & 17.17 & 20.83 & \textbf{25.85} \\
Direct Replay & 59.14 & 57.47 & 8.82 & 3.85 & 25.25 & 25.00 & \textbf{29.92} \\
Replay + Exploration (Full MM) & 80.65 & 86.21 & 49.02 & 7.69 & 34.34 & 31.25 & \textbf{48.19} \\
\bottomrule
\end{tabular}
\end{table}

\textbf{Replay improves the effectiveness of tutorial-derived skills.}
Using the extracted candidates directly raises mean task success from 19.92\% (Vanilla) to 25.85\% (Candidate). Much of this gain comes from KAlgebra, while Candidate scores below Vanilla on GRASS GIS, Celestia, and TeXstudio.
Direct Replay refines the procedures through execution in the GUI and retains those that achieve their intended effects. Using these verified skills and their execution images raises mean success to 29.92\%. The largest gains over Candidate are on Celestia (17.17\% to 25.25\%) and ChimeraX (49.43\% to 57.47\%).

\textbf{Exploration yields substantial gains by expanding the verified skill library beyond the original tutorials.}
Adding related operations that pass Replay grows the library from 467 to 997 verified skills. In Table~\ref{tab:stage_ablation}, mean task success rises from 29.92\% to 48.19\%. All six applications improve, with the largest gain on GRASS GIS (8.82\% to 49.02\%).

\subsubsection{Knowledge Type Ablation}
\label{sec:knowledge_type_ablation}
Scientific GUI tasks draw on two kinds of knowledge. Knowledge Cards explain domain concepts and application behavior, such as why orbital data become outdated in Celestia. Skills provide procedural knowledge for GUI operations, such as the steps for fetching a molecular structure by ID in ChimeraX.
To measure what each contributes, we keep the final library fixed and allow GPT-5.5 to retrieve Cards alone, skills alone with their available reference images, or both (Table~\ref{tab:knowledge_type_ablation}).

\begin{table}[!t]
\centering
\caption{Knowledge type ablation with GPT-5.5. Task success (\%) is measured with Knowledge Cards, verified skills from the final library, or both. Vanilla uses neither.}
\label{tab:knowledge_type_ablation}
\small
\setlength{\tabcolsep}{4pt}
\renewcommand{\arraystretch}{1.10}
\begin{tabular}{lrrrrrrr}
\toprule
Resources at execution & Alg & Biochem & GIS & ATP & Astron & Doc & \textbf{Avg} \\
\midrule
Vanilla & 12.90 & 44.83 & 12.75 & 3.85 & 20.20 & 25.00 & \textbf{19.92} \\
Card-only & 48.39 & 65.52 & 31.37 & 3.85 & 20.20 & 25.00 & \textbf{32.39} \\
Skill-only (MM) & 64.52 & 86.21 & 42.16 & 5.13 & 21.21 & 27.08 & \textbf{41.05} \\
Cards + skills (Full MM) & 80.65 & 86.21 & 49.02 & 7.69 & 34.34 & 31.25 & \textbf{48.19} \\
\bottomrule
\end{tabular}
\end{table}

Both types of knowledge improve on the Vanilla mean of 19.92\%. Cards alone reach 32.39\% mean success and skills alone reach 41.05\%. Together they reach 48.19\%.
The added value of Cards depends on the application. On KAlgebra, supplying Cards alongside skills raises success from 64.52\% to 80.65\%. On ChimeraX, skills alone already match the combined result of 86.21\%.

\subsubsection{Number of Exploration Rounds}
\label{sec:continued_exploration}
\begin{figure}[!t]
\centering
\begin{tikzpicture}
\definecolor{skillred}{HTML}{D9454C}
\definecolor{successpurple}{HTML}{3D1E96}
\pgfplotstableread{
round skills success
0 467 29.92
1 567 35.51
2 667 36.47
3 757 39.42
4 837 42.95
5 897 45.88
6 957 48.19
7 997 48.19
8 997 48.19
}\rounddata
\begin{groupplot}[
  group style={group size=1 by 2, vertical sep=27pt},
  scale only axis, width=0.86\linewidth, height=47pt,
  xmin=-0.35, xmax=8.35, xtick={0,...,8},
  axis lines*=left, axis line style={black!45, thin},
  tick style={black!45}, tick align=outside, tickwidth=2pt,
  ymajorgrids, grid style={black!12, thin},
  tick label style={font=\footnotesize},
  title style={at={(axis description cs:0,1)},anchor=south west,
               xshift=-3pt,yshift=1pt,font=\small\bfseries},
  point meta=y,
  every node near coord/.append style={font=\footnotesize,anchor=south,
                                      inner sep=1pt,yshift=2pt},
  clip=false]
\nextgroupplot[title={Total verified skills}, ymin=400, ymax=1150,
              ytick={500,700,900}, xticklabels=\empty]
\addplot[skillred, line width=1.15pt, mark=square*, mark size=2pt,
         nodes near coords={\pgfmathprintnumber[fixed,precision=0]{\pgfplotspointmeta}}]
  table[x=round,y=skills]{\rounddata};
\nextgroupplot[title={GPT-5.5 task success (\%)}, ymin=27, ymax=55,
              ytick={30,40,50}, xticklabels={R0,R1,R2,R3,R4,R5,R6,R7,R8},
              xlabel={Exploration round}, xlabel style={font=\small}]
\addplot[successpurple, line width=1.15pt, mark=*, mark size=2pt,
         nodes near coords={\pgfmathprintnumber[fixed,fixed zerofill,precision=2]{\pgfplotspointmeta}\%}]
  table[x=round,y=success]{\rounddata};
\end{groupplot}
\end{tikzpicture}
\caption{Exploration on six ScienceBoard applications with GPT-5.5. The upper panel shows the total number of verified skills across applications. The lower panel shows the mean task success rate, weighting the six applications equally. R0 denotes Direct Replay.}
\label{fig:exploration_rounds}
\end{figure}
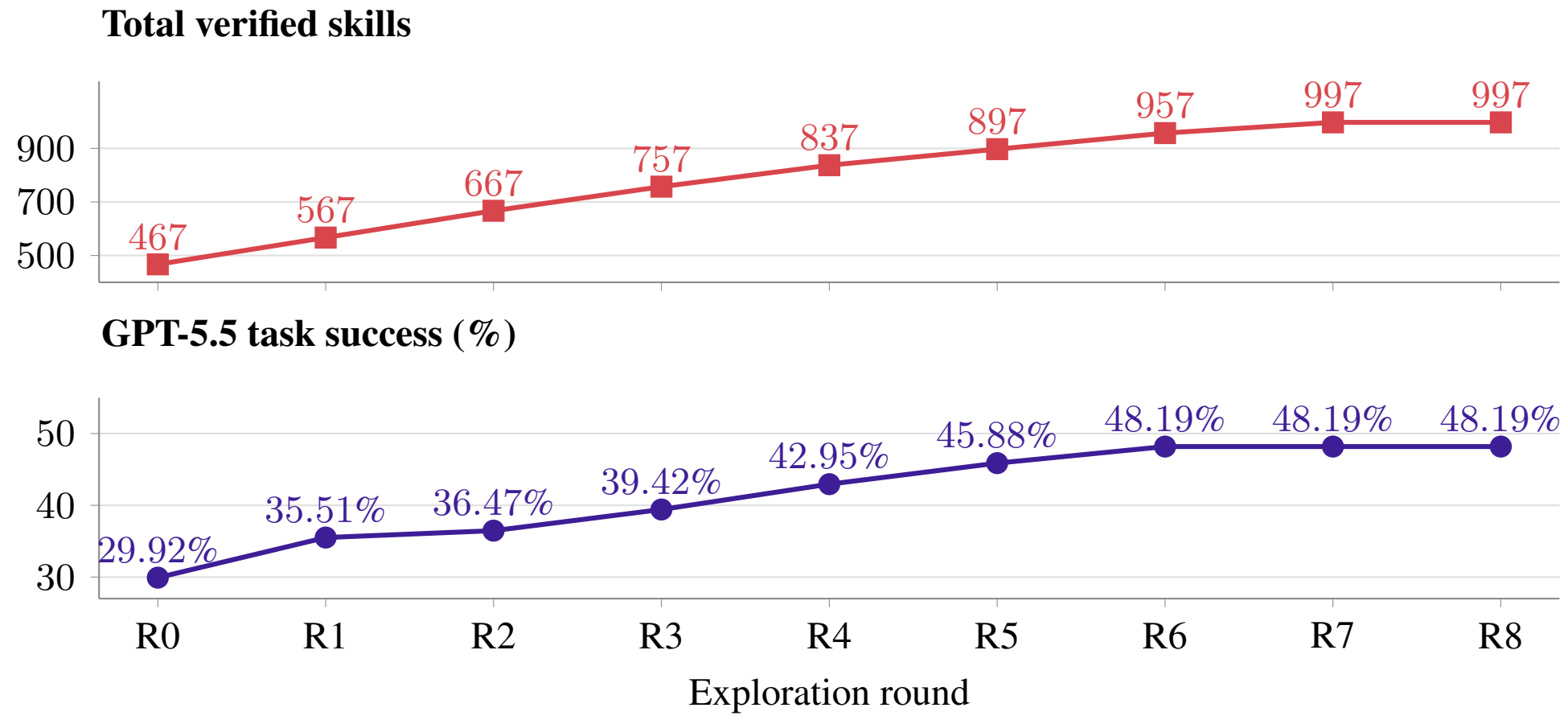

Figure~\ref{fig:exploration_rounds} tracks library size and downstream task success as we expand the skills available to GPT-5.5 in the multimodal setting. Starting from Direct Replay (R0), we run up to eight rounds of Exploration, stopping for each application when a round yields no new verified skill.

Across the six applications, the library grows from 467 verified skills at R0 to 957 at R6, and mean task success rises in every round, from 29.92\% to 48.19\%. Mean success then holds at 48.19\% through R8. The library continues to grow for one more round, reaching 997 skills at R7 before expansion yields no new verified skills at R8. Appendix~\ref{app:exploration_by_app} gives the success rates and skill counts for each application and round (Tables~\ref{tab:continued_exploration} and~\ref{tab:task_skill_counts}).

\section{Conclusion}
\label{sec:conclusion}

TutorialGUI uses interaction with the target software to verify procedures extracted from tutorials and to explore related operations. GUI agents can then retrieve verified skills alongside Knowledge Cards that explain application concepts and behavior. On 169 tasks across six ScienceBoard applications, three executors reused the same frozen library without fine-tuning. With visual references, their task success improved by an average of 19.45 percentage points over their respective Vanilla settings. For GPT-5.5, mean success on these tasks rose from 29.92\% after Direct Replay to 48.19\% by the sixth round of Exploration and then leveled off, one round before the skill library reached its final size.

\section*{Limitations and Future Work}

Official tutorials may provide incomplete coverage of application capabilities, which can limit the scope of the initial resource library.
More extensive Exploration may therefore be needed for applications with sparse documentation.

The exploration study tracks one executor across six applications. On KAlgebra, additional skills after R5 yield the same average success on the evaluation tasks. Distinguishing overlap in the skills from limits in selection or execution would require a separate analysis.

\bibliography{iclr2027_conference}
\bibliographystyle{iclr2027_conference}

\clearpage
\appendix
\raggedbottom
\section{Supplementary Experimental Details and Prompts}

\subsection{Exploration Results by Application}
\label{app:exploration_by_app}
Tables~\ref{tab:continued_exploration} and~\ref{tab:task_skill_counts} give the application-level results behind Figure~\ref{fig:exploration_rounds}. The application columns appear in the same order in both tables. The last columns give the mean success rate and the total number of verified skills, respectively.

\begin{table}[H]
\centering
\caption{GPT-5.5 task success (\%) by application and Exploration round. Each application column is a three-run mean. Avg is the unweighted mean across applications.}
\label{tab:continued_exploration}
\small
\setlength{\tabcolsep}{4pt}
\renewcommand{\arraystretch}{1.10}
\begin{tabular}{@{}p{8.6em}*{7}{>{\raggedleft\arraybackslash}p{3.7em}}@{}}
\toprule
Exploration round & Alg & Biochem & GIS & ATP & Astron & Doc & \textbf{Avg} \\
\midrule
R0 (Direct Replay) & 59.14 & 57.47 & 8.82 & 3.85 & 25.25 & 25.00 & \textbf{29.92} \\
R1 & 64.52 & 66.67 & 22.55 & 3.85 & 24.24 & 31.25 & \textbf{35.51} \\
R2 & 65.59 & 67.82 & 23.53 & 6.41 & 24.24 & 31.25 & \textbf{36.47} \\
R3 & 72.04 & 74.71 & 23.53 & 7.69 & 27.27 & 31.25 & \textbf{39.42} \\
R4 & 76.34 & 81.61 & 26.47 & 7.69 & 34.34 & 31.25 & \textbf{42.95} \\
R5 & 80.65 & 85.06 & 36.27 & 7.69 & 34.34 & 31.25 & \textbf{45.88} \\
R6 & 80.65 & 86.21 & 49.02 & 7.69 & 34.34 & 31.25 & \textbf{48.19} \\
R7 & 80.65 & 86.21 & 49.02 & 7.69 & 34.34 & 31.25 & \textbf{48.19} \\
R8 & 80.65 & 86.21 & 49.02 & 7.69 & 34.34 & 31.25 & \textbf{48.19} \\
\bottomrule
\end{tabular}
\end{table}

KAlgebra reaches its final success rate of 80.65\% at R5 with 195 skills, while GRASS GIS improves from 36.27\% at R5 to 49.02\% at R6. Only KAlgebra and ChimeraX add skills at R7 (20 each). TeXstudio reaches 55 skills at R2. Its last round, R3, adds no skills, and the executor solves 5 of 16 tasks in each of the three runs (31.25\%).

\begin{table}[H]
\centering
\caption{Verified skill counts by application and Exploration round. Target is the number of newly verified skills sought per round while an application is still expanding. Counts are cumulative from R0.}
\label{tab:task_skill_counts}
\small
\setlength{\tabcolsep}{4pt}
\renewcommand{\arraystretch}{1.10}
\begin{tabular}{@{}p{8.6em}*{7}{>{\raggedleft\arraybackslash}p{3.7em}}@{}}
\toprule
Library measure & Alg & Biochem & GIS & ATP & Astron & Doc & \textbf{Total} \\
\midrule
Evaluation tasks & 31 & 29 & 34 & 26 & 33 & 16 & \textbf{169} \\
Target per round & 20 & 20 & 20 & 10 & 20 & 10 & --- \\
\midrule
R0 (Direct Replay) & 95 & 107 & 77 & 57 & 96 & 35 & \textbf{467} \\
R1 & 115 & 127 & 97 & 67 & 116 & 45 & \textbf{567} \\
R2 & 135 & 147 & 117 & 77 & 136 & 55 & \textbf{667} \\
R3 & 155 & 167 & 137 & 87 & 156 & 55 & \textbf{757} \\
R4 & 175 & 187 & 157 & 87 & 176 & 55 & \textbf{837} \\
R5 & 195 & 207 & 177 & 87 & 176 & 55 & \textbf{897} \\
R6 & 215 & 227 & 197 & 87 & 176 & 55 & \textbf{957} \\
R7 & 235 & 247 & 197 & 87 & 176 & 55 & \textbf{997} \\
R8 & 235 & 247 & 197 & 87 & 176 & 55 & \textbf{997} \\
\bottomrule
\end{tabular}
\end{table}

The expanded library contains 997 verified skills and is used alongside 1,715 Knowledge Cards. The parameter-change experiment in Appendix~\ref{app:parameter_verification} uses a separate fixed collection of 751 skills, of which 349 explicitly declare parameters.

\clearpage
\subsection{Experimental Environment and Implementation Details}
\label{app:experimental_details}

\paragraph{Environment and Software.}
Replay and downstream evaluation are conducted in AgentBay sandboxes running Ubuntu 22.04, each configured with 8 vCPUs, 16 GiB of memory, and a fixed screen resolution of $1920 \times 1080$.
We follow the application versions specified by ScienceBoard~\citep{sun2026scienceboard}.
Table~\ref{tab:software_versions} maps the domain abbreviations used in Table~\ref{tab:main_results} to the corresponding applications and software configurations in our sandboxes.

\begin{table}[H]
    \centering
    \caption{
    Application configurations in our AgentBay sandboxes for Replay and downstream evaluation.
    Software versions follow ScienceBoard~\citep{sun2026scienceboard}, and the rows correspond to the domain columns in Table~\ref{tab:main_results}.
    }
    \label{tab:software_versions}
    \begin{tabular}{lll}
        \toprule
        Domain & Application & Configuration \\
        \midrule
        Alg     & KAlgebra              & 21.12.3 \\
        Biochem & ChimeraX              & 1.9 \\
        GIS     & GRASS GIS             & 8.4.2 \\
        ATP     & Lean 4 within VS Code & VS Code 1.127.0; Lean 4 extension 0.0.237 \\
        Astron  & Celestia              & 1.6.4 \\
        Doc     & TeXstudio             & 4.8.6 \\
        \bottomrule
    \end{tabular}
\end{table}

\paragraph{Resource Construction.}
\label{app:implementation_details}
A unified model configuration is used across extraction, Replay, and Exploration.
Candidate-skill and Knowledge Card extraction, parameter instantiation and action planning, visual outcome verification, and candidate generation during Exploration are performed with Qwen3.8-Max.
Action grounding is handled by GUI-Plus%
\footnote{\url{https://help.aliyun.com/en/model-studio/gui-automation}},
which is specialized for GUI understanding and interaction.
Candidates proposed during Exploration are subsequently executed and verified through Replay before entering the skill library.

\textit{Replay.}
Each candidate is allowed at most three Replay attempts, including the initial attempt.
Replay terminates once verification succeeds.
Candidates that do not pass verification within this budget are discarded and are not retried in subsequent Exploration rounds.

\textit{Exploration.}
Each candidate-generation call has its own proposal budget. Within a round, we alternate proposal calls with Replay and count each newly verified skill when it enters the library. Reaching the application-specific target completes the round. If further proposals yield no new verified skills and the target can no longer be met, expansion ends for that application. Table~\ref{tab:task_skill_counts} lists the targets and cumulative library sizes.

\paragraph{Construction cost.}
Table~\ref{tab:construction_cost} reports token use by application. We estimate each application's Exploration total by multiplying its one-round token cost by the number of rounds run, including the final round that produces no new verified skills. For each application in Table~\ref{tab:task_skill_counts}, this is the first round in which its skill count stays the same.

\begin{table}[H]
\centering
\caption{Skill library construction costs by application, in millions of input and output tokens combined. Exploration totals are estimates.}
\label{tab:construction_cost}
\small
\setlength{\tabcolsep}{7pt}
\renewcommand{\arraystretch}{1.10}
\begin{tabular}{lrr}
\toprule
Application & Direct Replay total & Exploration total \\
\midrule
KAlgebra & 8.529 & 5.696 \\
ChimeraX & 33.978 & 2.904 \\
GRASS GIS & 20.702 & 5.425 \\
Lean 4 & 11.009 & 2.036 \\
Celestia & 19.556 & 1.600 \\
TeXstudio & 10.058 & 1.683 \\
\midrule
Total & 103.832 & 19.344 \\
\bottomrule
\end{tabular}
\end{table}

\paragraph{Downstream Evaluation.}
After resource construction, the skill library is frozen and shared across all evaluated models.
TutorialGUI is evaluated in the AgentBay environments described above, using the official ScienceBoard task-success evaluators~\citep{sun2026scienceboard}.
GPT-5.5, Kimi-K2.5, and Qwen3.8-Max each serve as both Selector and Executor following the resource-selection protocol in Section~\ref{sec:skill_augmented_execution}.

\textit{TutorialGUI-Text} provides the selected Knowledge Cards and skills in textual form.
\textit{TutorialGUI-MM} follows the same protocol and additionally provides a reference image for a selected skill when available.
Complete abstention provides no external image.
Both settings retain live screenshots of the current GUI state throughout task execution.
All model parameters remain fixed during resource construction and downstream evaluation.
Appendix~\ref{app:comparison_protocol} describes our Vanilla runs and the published results for trained GUI agents.

\subsection{Tutorial Sources}
\label{app:tutorial_sources}

Table~\ref{tab:tutorial_sources} lists the tutorial and documentation sources used for knowledge extraction, together with their resource formats. For the Celestia guides page, MS Word denotes the linked guide documents rather than the format of the landing page.

\begin{table}[H]
    \centering
    \caption{Tutorial and documentation sources by application. URLs identify the source pages or repositories; formats describe the corresponding resources.}
    \label{tab:tutorial_sources}
    \small
    \renewcommand{\arraystretch}{1.2}
    \begin{tabular}{@{}p{0.16\linewidth}p{0.64\linewidth}p{0.12\linewidth}@{}}
        \toprule
        Application & Source URL & Format \\
        \midrule
        KAlgebra & \url{https://docs.kde.org/stable_kf6/en/kalgebra/kalgebra/} & HTML \\
        \addlinespace
        ChimeraX & \url{https://www.cgl.ucsf.edu/chimerax/docs/user/} & HTML \\
        \addlinespace
        GRASS GIS & \url{https://grass.osgeo.org/grass84/manuals/} & HTML \\
        \addlinespace
        Lean 4 & \url{https://leanprover-community.github.io/mathematics_in_lean/} & HTML \\
        & \url{https://leanprover-community.github.io/learn.html} & HTML \\
        & \url{https://github.com/leanprover/vscode-lean4/blob/master/vscode-lean4/manual/manual.md} & Markdown \\
        \addlinespace
        Celestia & \url{https://en.wikibooks.org/wiki/Celestia} & HTML \\
        & \url{https://celestiaproject.space/guides.html} & MS Word \\
        \addlinespace
        TeXstudio & \url{https://texstudio-org.github.io} & HTML \\
        \bottomrule
    \end{tabular}
\end{table}

\subsection{Comparison Methods and Reporting Conventions}
\label{app:comparison_protocol}

\paragraph{Vanilla runs.}
We ran each of the three models in the Vanilla setting on all 169 tasks in the same AgentBay environments used to evaluate TutorialGUI, with no TutorialGUI Knowledge Cards or skills. We scored each run with the official ScienceBoard evaluators and report the mean over three runs per task in Table~\ref{tab:main_results}.

\paragraph{Published comparison results.}
CODA Stage-2 results are the Average@8 scores from Table 1 of \citet{sun2025coda}, using a trained Qwen2.5-VL-32B planner and a UI-TARS-1.5-7B executor. SEAgent results are the UI-TARS-1.5-7B-DPO+SEAgent success rates from Table 5 of \citet{sun2025seagent}. ScaleCUA results are the screenshot-observation Qwen3.5-9B scores from Table 3 of \citet{lv2026scalecua}, which reports four independent rollout passes per task. These published results retain their original backbone, training, and evaluation settings.

\paragraph{Aggregation and performance gains.}
For our evaluations, application-level success rates are averaged over three runs. Avg gives equal weight to each reported application: four for CODA and SEAgent, and six for ScaleCUA and our evaluated models. Missing entries (---) are excluded. Published per-application scores retain their original precision; applying this averaging convention to ScaleCUA yields 55.23\%, compared with its reported Overall of 54.0\%. For our results, averages and performance differences are computed from unrounded success proportions and rounded to two decimal places only for presentation. In Table~\ref{tab:main_results}, $\Delta$ gives the difference in Avg from the corresponding Vanilla score in percentage points.

\Needspace{11\baselineskip}
\subsection{Run-to-Run Variability}
\label{app:run_variability}
For each application and run, we divide the number of successful tasks by the application's task count in Table~\ref{tab:task_skill_counts}. Each entry reports the mean of the three success rates (\%) and their sample standard deviation (pp), using $n-1=2$ in the variance denominator. For Avg, we first average the six application rates within each run, giving equal weight to each application. We then compute the mean and sample standard deviation of the three run averages and round only the reported values.

\begin{table}[H]
\centering
\caption{Cross-model results for Vanilla, MMG2Skill, and both TutorialGUI variants in Table~\ref{tab:main_results}, summarized across three runs.}
\label{tab:main_run_variability}
\small
\setlength{\tabcolsep}{5pt}
\renewcommand{\arraystretch}{1.10}
\begin{tabular}{lrrrr}
\toprule
Application & Vanilla & MMG2Skill & TutorialGUI-Text & TutorialGUI-MM \\
\midrule
\rowcolor[gray]{0.92}
\multicolumn{5}{l}{\textit{GPT-5.5}} \\
KAlgebra & $12.90 \pm 3.23$ & $48.39 \pm 3.23$ & $67.74 \pm 6.45$ & $80.65 \pm 3.23$ \\
ChimeraX & $44.83 \pm 3.45$ & $50.57 \pm 1.99$ & $74.71 \pm 1.99$ & $86.21 \pm 0.00$ \\
GRASS GIS & $12.75 \pm 1.70$ & $22.55 \pm 1.70$ & $40.20 \pm 1.70$ & $49.02 \pm 4.49$ \\
Lean 4 & $3.85 \pm 0.00$ & $3.85 \pm 0.00$ & $5.13 \pm 2.22$ & $7.69 \pm 0.00$ \\
Celestia & $20.20 \pm 1.75$ & $19.19 \pm 1.75$ & $30.30 \pm 3.03$ & $34.34 \pm 1.75$ \\
TeXstudio & $25.00 \pm 0.00$ & $4.17 \pm 3.61$ & $27.08 \pm 3.61$ & $31.25 \pm 0.00$ \\
Avg & $19.92 \pm 0.48$ & $24.79 \pm 0.58$ & $40.86 \pm 1.51$ & $48.19 \pm 1.02$ \\
\midrule
\rowcolor[gray]{0.92}
\multicolumn{5}{l}{\textit{Kimi-K2.5}} \\
KAlgebra & $10.75 \pm 1.86$ & $32.26 \pm 3.23$ & $29.03 \pm 3.23$ & $29.03 \pm 3.23$ \\
ChimeraX & $29.89 \pm 1.99$ & $52.87 \pm 5.27$ & $40.23 \pm 1.99$ & $52.87 \pm 1.99$ \\
GRASS GIS & $0.98 \pm 1.70$ & $9.80 \pm 1.70$ & $14.71 \pm 0.00$ & $11.76 \pm 0.00$ \\
Lean 4 & $0.00 \pm 0.00$ & $3.85 \pm 0.00$ & $7.69 \pm 3.85$ & $3.85 \pm 0.00$ \\
Celestia & $5.05 \pm 1.75$ & $22.22 \pm 1.75$ & $18.18 \pm 0.00$ & $23.23 \pm 1.75$ \\
TeXstudio & $8.33 \pm 3.61$ & $12.50 \pm 0.00$ & $20.83 \pm 3.61$ & $25.00 \pm 0.00$ \\
Avg & $9.17 \pm 0.77$ & $22.25 \pm 1.95$ & $21.78 \pm 0.72$ & $24.29 \pm 1.08$ \\
\midrule
\rowcolor[gray]{0.92}
\multicolumn{5}{l}{\textit{Qwen3.8-Max}} \\
KAlgebra & $41.94 \pm 3.23$ & $55.91 \pm 1.86$ & $74.19 \pm 0.00$ & $74.19 \pm 3.23$ \\
ChimeraX & $60.92 \pm 3.98$ & $58.62 \pm 3.45$ & $78.16 \pm 7.18$ & $82.76 \pm 3.45$ \\
GRASS GIS & $17.65 \pm 0.00$ & $17.65 \pm 2.94$ & $14.71 \pm 2.94$ & $14.71 \pm 2.94$ \\
Lean 4 & $10.26 \pm 2.22$ & $11.54 \pm 3.85$ & $12.82 \pm 2.22$ & $8.97 \pm 2.22$ \\
Celestia & $22.22 \pm 1.75$ & $31.31 \pm 1.75$ & $38.38 \pm 1.75$ & $45.45 \pm 3.03$ \\
TeXstudio & $29.17 \pm 7.22$ & $37.50 \pm 0.00$ & $31.25 \pm 6.25$ & $45.83 \pm 3.61$ \\
Avg & $30.36 \pm 2.35$ & $35.42 \pm 0.27$ & $41.59 \pm 0.65$ & $45.32 \pm 0.65$ \\
\bottomrule
\end{tabular}
\end{table}

\begin{table}[H]
\centering
\caption{Variability for the three additional GPT-5.5 comparison methods in Table~\ref{tab:main_results}.}
\label{tab:comparison_variability}
\small
\setlength{\tabcolsep}{4pt}
\renewcommand{\arraystretch}{1.10}
\begin{tabular}{lrrr}
\toprule
Application & Tutorial RAG & VisualSkill & Resource2Skill \\
\midrule
KAlgebra & $40.86 \pm 6.72$ & $52.69 \pm 11.33$ & $29.03 \pm 0.00$ \\
ChimeraX & $51.72 \pm 3.45$ & $58.62 \pm 3.45$ & $56.32 \pm 5.27$ \\
GRASS GIS & $8.82 \pm 2.94$ & $17.65 \pm 0.00$ & $23.53 \pm 2.94$ \\
Lean 4 & $3.85 \pm 0.00$ & $5.13 \pm 2.22$ & $3.85 \pm 0.00$ \\
Celestia & $21.21 \pm 3.03$ & $27.27 \pm 3.03$ & $25.25 \pm 4.63$ \\
TeXstudio & $20.83 \pm 3.61$ & $10.42 \pm 3.61$ & $10.42 \pm 3.61$ \\
\midrule
Avg & $24.55 \pm 1.72$ & $28.63 \pm 1.56$ & $24.73 \pm 2.43$ \\
\bottomrule
\end{tabular}
\end{table}

\begin{table}[H]
\centering
\caption{Construction and resource ablations with GPT-5.5. Vanilla and Full (MM) are reported in Table~\ref{tab:main_run_variability}.}
\label{tab:stage_run_variability}
\small
\setlength{\tabcolsep}{4pt}
\renewcommand{\arraystretch}{1.10}
\begin{tabular}{lrrrr}
\toprule
Application & Candidate & Direct Replay & Card-only & Skill-only \\
\midrule
KAlgebra & $56.99 \pm 8.12$ & $59.14 \pm 1.86$ & $48.39 \pm 0.00$ & $64.52 \pm 3.23$ \\
ChimeraX & $49.43 \pm 7.18$ & $57.47 \pm 5.27$ & $65.52 \pm 0.00$ & $86.21 \pm 0.00$ \\
GRASS GIS & $6.86 \pm 1.70$ & $8.82 \pm 2.94$ & $31.37 \pm 1.70$ & $42.16 \pm 1.70$ \\
Lean 4 & $3.85 \pm 0.00$ & $3.85 \pm 0.00$ & $3.85 \pm 0.00$ & $5.13 \pm 2.22$ \\
Celestia & $17.17 \pm 3.50$ & $25.25 \pm 1.75$ & $20.20 \pm 1.75$ & $21.21 \pm 3.03$ \\
TeXstudio & $20.83 \pm 3.61$ & $25.00 \pm 0.00$ & $25.00 \pm 6.25$ & $27.08 \pm 3.61$ \\
\midrule
Avg & $25.85 \pm 1.45$ & $29.92 \pm 0.85$ & $32.39 \pm 1.30$ & $41.05 \pm 1.61$ \\
\bottomrule
\end{tabular}
\end{table}

\Needspace{16\baselineskip}
\subsection{Verification under Parameter Changes}
\label{app:parameter_verification}
For this study we used a library of 751 skills across six applications and tested all 349 skills that explicitly declare at least one parameter. For each skill, we executed its procedure with four distinct new inputs and checked its declared postconditions for each new parameter binding. We compared each resulting verification label with the label recorded for the original input. The labels matched in 1,364 of the 1,396 new-input executions, yielding a pooled agreement of 97.71\% (Table~\ref{tab:parameter_verification}).

\begin{table}[H]
\centering
\caption{Verification under parameter changes. Each parameterized skill contributes one original input and four new inputs. Agreement is the percentage of new-input cases whose verification label matches the original label.}
\label{tab:parameter_verification}
\small
\setlength{\tabcolsep}{4pt}
\renewcommand{\arraystretch}{1.10}
\begin{tabular}{@{}lrrrrrr@{}}
\toprule
 & \multicolumn{2}{c}{Skills} & \multicolumn{2}{c}{Input cases} & \multicolumn{2}{c}{New-input results} \\
\cmidrule(lr){2-3}\cmidrule(lr){4-5}\cmidrule(l){6-7}
Application & All & Parameterized & All & New & Matches & Agreement (\%) \\
\midrule
KAlgebra & 170 & 56 & 280 & 224 & 212 & 94.64 \\
ChimeraX & 170 & 96 & 480 & 384 & 377 & 98.18 \\
GRASS GIS & 149 & 145 & 725 & 580 & 572 & 98.62 \\
Lean 4 & 90 & 19 & 95 & 76 & 75 & 98.68 \\
Celestia & 135 & 12 & 60 & 48 & 46 & 95.83 \\
TeXstudio & 37 & 21 & 105 & 84 & 82 & 97.62 \\
\midrule
Total & 751 & 349 & 1,745 & 1,396 & 1,364 & 97.71 \\
\bottomrule
\end{tabular}
\end{table}

\Needspace{11\baselineskip}
\begingroup
\let\promptquote\quote
\renewcommand{\quote}{\promptquote\small\setlength{\parskip}{2pt plus 0.5pt}}
\subsection{Key Prompts for Resource Construction and Task Execution}
\label{app:key_prompts}

The prompts below retain the core instructions, with output schemas and formatting requirements omitted for brevity.

\begingroup
\raggedbottom

\subsubsection{Source-Grounded Knowledge Extraction}

Knowledge extraction produces source-grounded Skill candidates, Knowledge Cards, and Unknown Slots, and may yield multiple records. Knowledge Cards provide non-executable decision support, while Unknown Slots identify information requiring further evidence rather than guessed values.

\begin{quote}
Mine the following approved source units for App \texttt{\{app\}}. Produce \texttt{\{skill\_count\}} distinct, replayable, low-risk Skill candidates when the evidence supports them; never pad the count.

A Skill must have one primary effect, semantic rather than coordinate targets, a required postcondition verifiable from exact application/file evidence or stable UI-semantic/declared visual-relation evidence, and no unresolved required field. One Skill may use one or several primitive actions. Frequently reused hotkey, menu, and button effects may stand alone when independently useful and verifiable.

Put variable values in typed runtime parameter slots instead of cloning literal-specific Skills. Exclude command-line-only actions unless the source explicitly presents them as an operation through the GUI.

Also extract source-supported, nonredundant Knowledge Cards when they contribute useful decision-relevant knowledge. Do not impose a fixed card count or generate cards merely to fill a quota. Return an empty card list when the sources contain no suitable knowledge.

A Knowledge Card is a non-executable knowledge record. It may explain when an operation is appropriate, how to interpret a parameter or application state, which constraints or prerequisites apply, how to recognize a valid result, or how to diagnose a failure. State a clear claim and preserve its relevant conditions, qualifiers, and exclusions. Explain which decision the knowledge supports. Do not duplicate an entire Skill procedure as a Card merely to create another record, and do not label a Card as a replay-verified executable Skill.

Return genuine Unknown Slots for information the sources do not establish. Do not invent missing facts. Candidates with unresolved required information must not be presented as replay-ready Skills.

Paraphrase rather than copying long source passages. Every extracted record must cite valid supplied source-unit IDs.

\end{quote}

\subsubsection{Replay Postcondition Verification Prompt}
\label{app:replay_verifier}

The verifier assesses declared postconditions for a single Replay attempt. The fixture contract and transient screenshots are used only when supplied and applicable, as specified below.

\begin{quote}
You are an independent postcondition verifier for one source-bound Atomic Skill Replay in a frozen desktop application.

Use only the declared postconditions, bound parameters, setup trace, action trace, stable before screenshot, current stable post-action screenshot, and any explicitly labelled immediate post-action transient full screenshot. The declared postconditions and actually bound parameters are authoritative. Planner or grounder prose cannot add a residue identity, object name, value, chain, target, or other success constraint that is absent from those declarations; when an optional parameter is unbound, a visibly valid runtime exemplar is acceptable if it satisfies the declared relation. Do not trust planner or grounder claims of completion. Do not use benchmark task IDs, answers, evaluator verdicts, historical pass/fail, or external facts. Evaluate only visible UI/application relations; file probes supplied by the executor take precedence for file-state assertions.

A \texttt{fixture\_verification\_contract} is trusted only as a description of an attempt-local runner-generated fixture when its content SHA matches the \texttt{fixture\_create} setup trace. It may explain whether a visible before/after transition is consistent with that exact input, but it cannot replace live evidence, prove an assertion by itself, or become part of the reusable Skill.

For a source-declared transient UI effect such as an OSD or toast, a sharp immediate post-action full screenshot may establish the assertion when it directly and unambiguously displays the declared effect; the required stable after screenshot proves the application subsequently settled, but the transient text is not required to remain visible in that later frame.

Assess every declared postcondition exactly once, assigning pass, fail, or inconclusive and explaining the supporting evidence. Return \texttt{verified} only when every required assertion passes. Use \texttt{failed} for clear contradictory evidence and \texttt{inconclusive} for unavailable or ambiguous evidence.
\end{quote}

\subsubsection{Shared Knowledge-Selection Prompt}

The shared Selector selects up to two skills and one Knowledge Card from the library for the current task.

\begin{quote}
You select reusable tutorial knowledge for a GUI executor. This is selection only; parameter binding and execution are handled by the executor.

Select at most two Skills and at most one Card, with at most three records total. A Card is optional: select one only when it supplies relevant, compatible knowledge that helps the executor choose an operation, interpret or bind parameters, understand the current state, respect constraints, verify results, or handle a failure. Do not select a Card merely because its topic matches the task.

One useful partial operation is sufficient. Decompose the task into necessary operations; choose records that explain how to perform one or more of them in the relevant app. Do not require a candidate to solve the whole scientific task, guarantee success, contain the final answer, name the current object, or already know runtime values.

A source example's object names and values are replaceable parameters unless the source explicitly fixes them as preconditions. Unknown GUI state is checked by the executor; an explicit incompatible precondition is grounds for rejection.

Use the source procedure and coverage to decide applicability. Prefer the most direct operation for the requested outcome, then complementary steps. A Card containing the needed command or UI guidance can be more useful than an unrelated executable Skill, but its selection does not establish that the operation has been executed or verified.

Do not fill the quota with irrelevant records. Abstain only when none of the candidates supplies a useful compatible operation or knowledge. For cross-app tasks, select at most one record per owner app, except when an explicit application-specific policy replaces this rule.

Identify the required task stages and indicate which are covered by the selected records. Do not bind parameters during selection. The executor binds parameter values using the task instruction and current GUI observations. Required values must be resolved before executing the corresponding operation. Selection is not execution or task success.
\end{quote}

\subsubsection{TeXstudio Cross-Application Selection Policy}

When the TeXstudio cross-application branch is enabled, this policy replaces the shared rule of at most one record per owner app. It is not an unconditional addition for all tasks, and the global budget of at most two Skills and one Card remains unchanged.

\begin{quote}
For TeXstudio cross-app tasks, select only from the allowed owner apps. Complementary records may share an owner, including a Card and Skills from the same app.

The global limits remain at most two Skills and one Card, three distinct records total. No owner must be represented; do not substitute or add records to balance owners.

Report uncovered required stages explicitly. Owner membership does not establish stage coverage.
\end{quote}

\subsubsection{Execution-Time Knowledge Context}

This additional method context guides the GUI executor in using selected knowledge, historical procedures, and reference images. It is not the complete executor system prompt and does not replace the underlying action or coordinate protocol.

\begin{quote}
TutorialGUI method context: selected tutorial guidance and eligible Skill programs.

In the multimodal setting, each selected Skill is accompanied by one reference image when available. Source steps and reference images are historical examples, not current observations or answers. Follow compatible source operations; bind targets from the current task and live GUI.

Use Knowledge Cards as contextual guidance for decisions, parameters, constraints, state interpretation, and verification. A Card is not an executable program, and its claim alone does not establish the current GUI state or prove that an action succeeded. Check its conditions against current observations when applicable.

Missing values may be acquired through the GUI, but unresolved placeholders must never be executed. Partial coverage is valid; complete uncovered stages yourself. Check current preconditions and results.

Reference-image coordinates are not live coordinates. Follow the system action/coordinate protocol unchanged.
\end{quote}

\subsubsection{Exploration Candidate-Generation Prompt}
\label{app:exploration_prompt}

This prompt takes an anchor selected by the outer exploration loop and proposes candidates for subsequent Replay. The outer loop counts skills admitted after Replay toward the application-specific target. Reaching the target completes the round. If no further skill can be verified and admitted, expansion stops for that application.

\begin{quote}
You are exploring reusable operations in a desktop application.

Given a verified anchor skill, the existing skill library, current application observations, available initial-state setup, relevant tutorial or manual evidence, a per-call candidate budget, and the remaining target for newly verified skills in this round, propose exploration candidates in the following directions:

\textbf{FORWARD} --- Discover downstream operations that use an object produced by the anchor or a result state established by it. Identify the output or state that connects the anchor to the candidate.

\textbf{BACKWARD} --- Discover operations that establish an input or precondition required by the anchor. These may supply a missing preparation process or provide an alternative preparation path. An existing verified predecessor is not required. When the initial-state setup already satisfies the preconditions, explain whether further preparation exploration is useful.

\textbf{LATERAL} --- Discover operations that share an input, serve a related goal, or provide a neighboring function. These may include alternative operating modes or configurations within the same skill family. Explain what changes in the procedure, application state, or observable outcome, without assuming an execution dependency.

Requirements:
\begin{itemize}
    \item Associate every candidate with the supplied anchor ID.
    \item Specify its input slots, preconditions, procedure, and expected effect.
    \item Ground proposals in the supplied evidence and distinguish observed facts from hypotheses.
    \item Compare each candidate with existing skills. State whether it may represent a distinct operation, an alternative mode, a parameterized variant, or an already covered capability.
    \item Prioritize useful, evidence-supported candidates with checkable effects. Avoid redundant proposals that add no exploration value.
    \item Keep exploration relations separate from execution dependencies. Each candidate admitted after separate Replay and verification is added as a new skill record. Retain existing skill records unchanged; do not merge candidates into them or extend their parameter coverage.
    \item Do not infer dependencies from semantic similarity alone.
    \item Do not invent controls, sources, execution traces, or successful results.
    \item Do not use benchmark tasks, answers, or evaluation outcomes.
    \item Do not force coverage of all three directions. Explain any direction for which no useful candidate is supported.
    \item Do not execute or verify proposals. Candidates remain proposed and unverified pending separate Replay and verification.
    \item Stay within the supplied per-call candidate budget. Proposed candidates do not count toward the verified-skill target; only admission after separate Replay and verification counts. If the remaining target is zero, return no new candidates.
\end{itemize}

For each direction, report findings and supporting evidence. For each candidate, report its relation to the anchor and existing skills, evidence, uncertainties, and the checks required during Replay. Identify related existing skills and their supporting evidence.
\end{quote}

\endgroup

\endgroup

\clearpage
\section{Examples of Skills and Knowledge Cards}
\label{app:resource_examples}

The following examples complement Appendix~\ref{app:key_prompts} with a ChimeraX skill and a Celestia Knowledge Card.

\subsection{Skill Example: Fetching a Structure in ChimeraX}
\label{app:skill_example}

\begin{figure}[H]
    \centering
    \begin{minipage}[t]{0.49\linewidth}
        \centering
        \includegraphics[width=\linewidth]{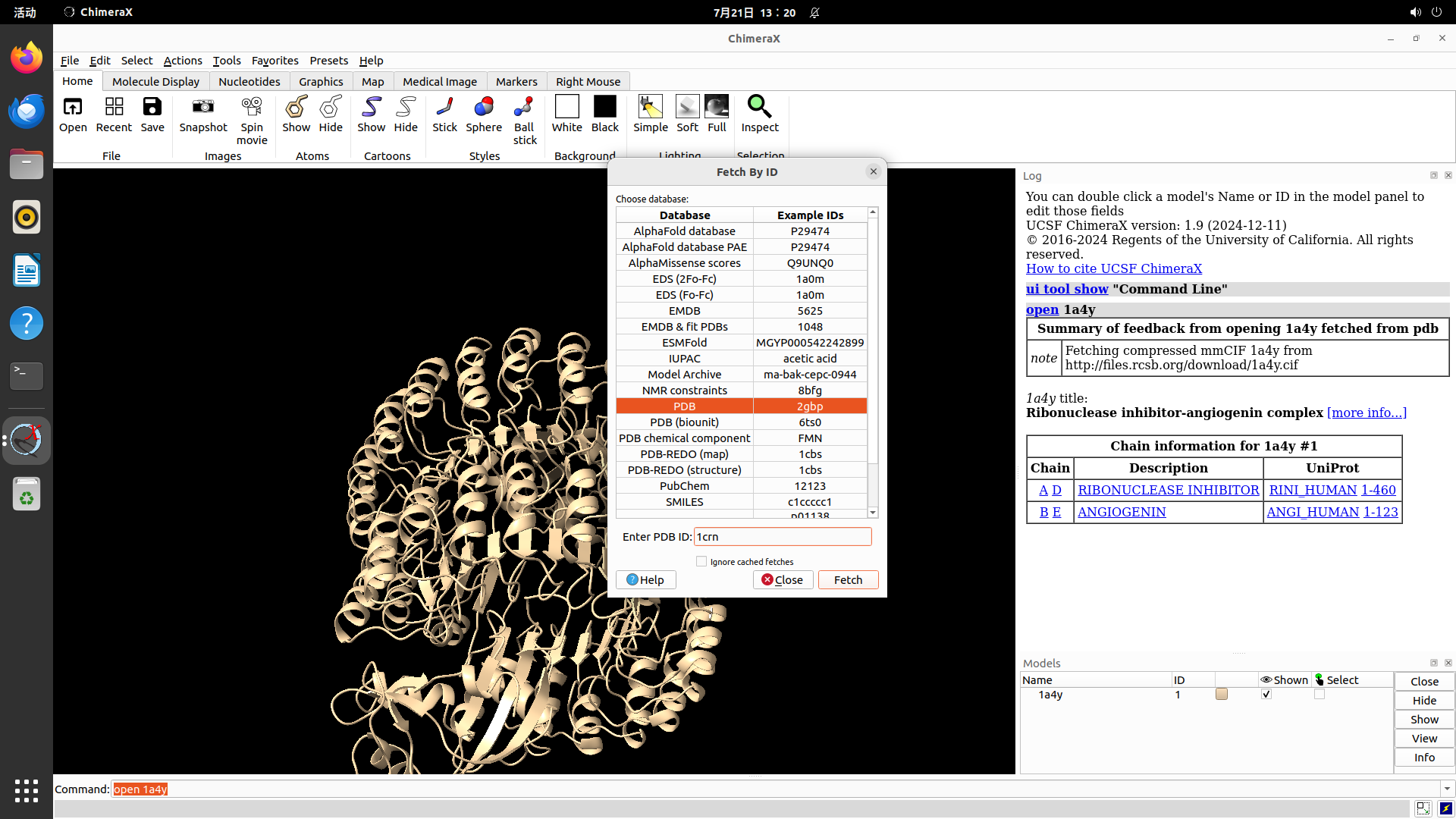}\par
        \vspace{4pt}\footnotesize\textbf{(a) Fetch by ID dialog}
    \end{minipage}\hfill
    \begin{minipage}[t]{0.49\linewidth}
        \centering
        \includegraphics[width=\linewidth]{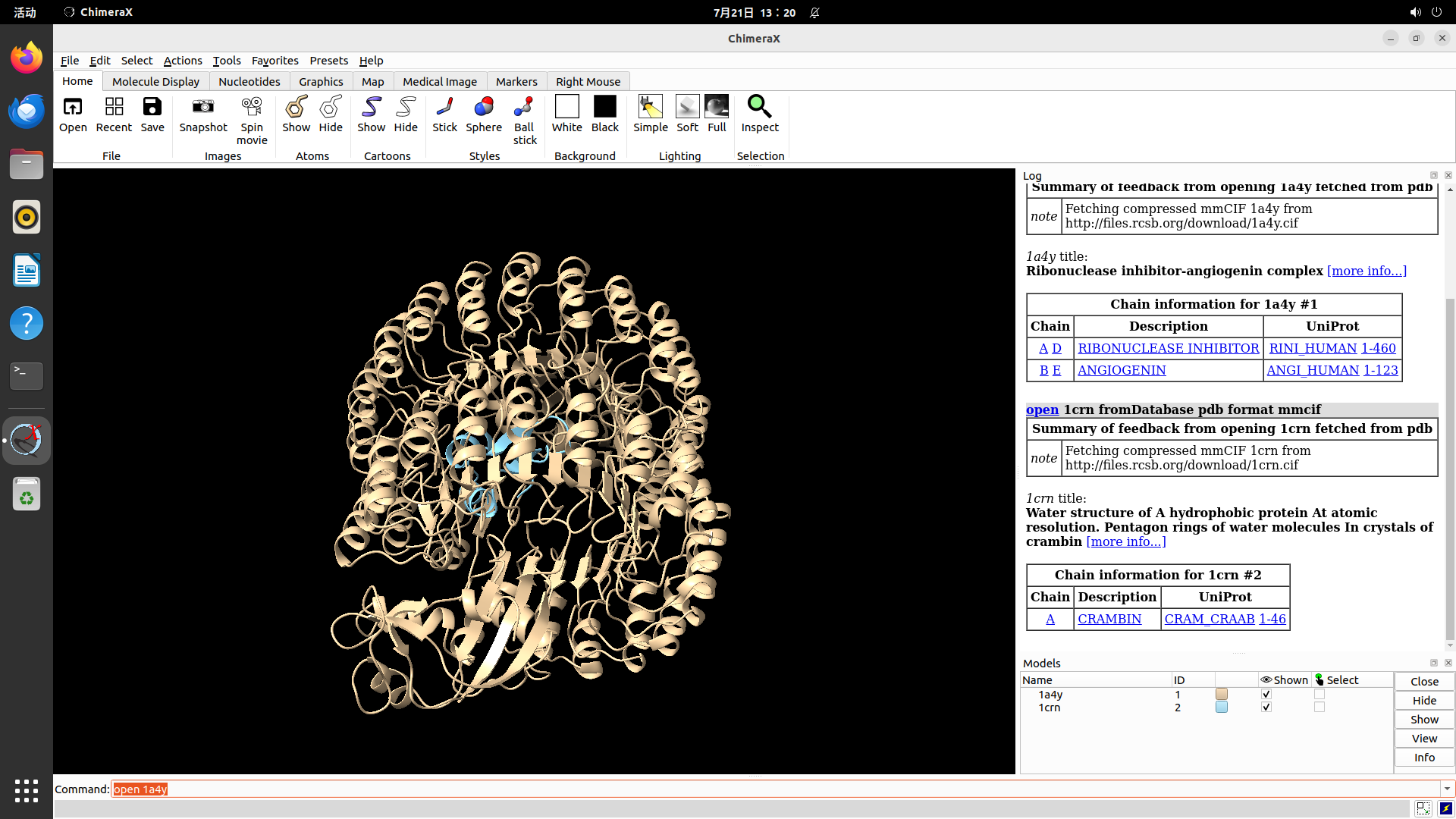}\par
        \vspace{4pt}\footnotesize\textbf{(b) Fetched model loaded}
    \end{minipage}
    \caption{Fetch by ID in ChimeraX. (a) PDB is selected and \texttt{1crn} entered. (b) Fetched \texttt{1CRN} (blue) appears alongside preparation model \texttt{1A4Y} (beige). The Models panel lists both entries, and the Log records the PDB fetch.}
    \label{fig:fetch-by-id-replay}
\end{figure}

Table~\ref{tab:skill_record_example} shows a parameterized Fetch by ID skill and its recorded verification results. Figure~\ref{fig:fetch-by-id-replay} shows the dialog and the fetched model.

\begin{table}[H]
    \centering
    \caption{Selected original fields of the ChimeraX Fetch by ID skill record.}
    \label{tab:skill_record_example}
    \small
    \renewcommand{\arraystretch}{1.16}
    \begin{tabular}{@{}p{0.25\linewidth}p{\dimexpr0.75\linewidth-2\tabcolsep\relax}@{}}
        \toprule
        Field & Value \\
        \midrule
        \texttt{name} & Fetch a structure through the Fetch by ID dialog \\
        \texttt{app} & ChimeraX \\
        \texttt{source} & \texttt{direct\_replay} \\
        \texttt{preconditions} & ChimeraX is running with an English, maximized main window. \\
        \texttt{input\_schema} & \texttt{database}, \texttt{identifier}: required strings \\
        \texttt{operation\_flow} & \texttt{open\_menu}: File\newline
        \texttt{click\_menu\_item}: Fetch by ID\newline
        \texttt{select\_control}: database\newline
        \texttt{type\_text}: identifier\newline
        \texttt{confirm}: Fetch \\
        \texttt{gui\_path} & \texttt{path}: [``File'', ``Fetch by ID'']\newline
        \texttt{visible\_result}: The requested database entry is fetched and opened as a model. \\
        \texttt{success\_checks} & The fetched model appears in the Model Panel and the Log identifies the completed fetch. \\
        \texttt{source\_citations} & \url{https://www.cgl.ucsf.edu/chimerax/docs/user/commands/open.html} \\
        \texttt{replay} & \texttt{fresh\_session}: \texttt{true}; \texttt{success}: \texttt{true} \\
        \texttt{semantic\_judgment} & \texttt{status}: \texttt{semantic\_pass}\newline
        \texttt{judge}: qwen3.8-max screenshot evaluator\newline
        \texttt{reason}: 1CRN loaded as new visible model (blue structure) while preparation model 1A4Y remains present (beige structure), both shown in Models panel \\
        \texttt{verified} & \texttt{true} \\
        \bottomrule
    \end{tabular}
\end{table}

\clearpage
\subsection{Knowledge Card Example: Understanding Outdated ISS Orbits in Celestia}
\label{app:card_example}

Table~\ref{tab:card_record_example} presents selected original fields of a Celestia Knowledge Card explaining why ISS orbit data can become outdated. This example illustrates how Knowledge Cards support diagnosis without specifying an executable procedure.
\begin{table}[H]
    \centering
    \caption{Selected original fields of the Celestia Knowledge Card record.}
    \label{tab:card_record_example}
    \small
    \renewcommand{\arraystretch}{1.15}
    \begin{tabular}{@{}p{0.30\linewidth}p{0.65\linewidth}@{}}
        \toprule
        Field & Value \\
        \midrule
        \texttt{app} & Celestia \\
        \texttt{artifact\_kind} & \texttt{knowledge\_card} \\
        \texttt{card\_type} & \texttt{troubleshooting} \\
        \texttt{content} & The orbit of the International Space Station changes continuously in ways that are almost impossible to predict due to things like atmospheric drag, light pressure, cargo ship docking, etc. \\
        \texttt{citation} & \texttt{source\_document\_id}: \texttt{wikibooks\_faqs}\newline
        \texttt{section\_id}: \texttt{wikibooks\_faqs:s00031}\newline
        \texttt{source\_locator}:\newline
        \url{https://en.wikibooks.org/wiki/Celestia/FAQs} \\
        \texttt{family\_ids} & [\texttt{celestia.time}] \\
        \texttt{capability\_tags} & [\texttt{celestia.time}, \texttt{celestia.time\_rate}] \\
        \texttt{runtime\_channel} & \texttt{phase0} \\
        \bottomrule
    \end{tabular}
\end{table}

\end{document}